\documentclass[10pt,twocolumn,letterpaper]{article}

\usepackage[pagenumbers]{cvpr} % To force page numbers, e.g. for an arXiv version

\definecolor{cvprblue}{rgb}{0.21,0.49,0.74}
\usepackage[pagebackref,breaklinks,colorlinks,allcolors=cvprblue]{hyperref}
\usepackage{multirow}

\usepackage{graphicx}
\usepackage{subcaption}
\usepackage{caption}
\usepackage{tikz}

\title{DynSUP: Dynamic Gaussian Splatting from An Unposed Image Pair}

\author{
Weihang Li\textsuperscript{1,3,*} \ \
Weirong Chen\textsuperscript{1,2,*} \ \
Shenhan Qian\textsuperscript{1,2} \ \
Jiajie Chen\textsuperscript{1,3} \ \
Daniel Cremers\textsuperscript{1,2} \ \
Haoang Li\textsuperscript{3} \\[3pt]
{\textsuperscript{1}Technical University of Munich}
\,
{\textsuperscript{2}Munich Center for Machine Learning}
\,\\
{\textsuperscript{3}The Hong Kong University of Science and Technology (Guangzhou)}
\\[3pt]
}

\begin{document}
\maketitle

\begin{abstract}
Recent advances in 3D Gaussian Splatting have shown promising results. Existing methods typically assume static scenes and/or multiple images with prior poses. Dynamics, sparse views, and unknown poses significantly increase the problem complexity due to insufficient geometric constraints. To overcome this challenge, we propose a method that can use only two images without prior poses to fit Gaussians in dynamic environments. To achieve this, we introduce two technical contributions. First, we propose an object-level two-view bundle adjustment. This strategy decomposes dynamic scenes into piece-wise rigid components, and jointly estimates the camera pose and motions of dynamic objects. Second, we design an SE(3) field-driven Gaussian training method. It enables fine-grained motion modeling through learnable per-Gaussian transformations. Our method leads to high-fidelity novel view synthesis of dynamic scenes while accurately preserving temporal consistency and object motion. Experiments on both synthetic and real-world datasets demonstrate that our method significantly outperforms state-of-the-art approaches designed for the cases of static environments, multiple images, and/or known poses. Our project page is
available at \url{https://colin-de.github.io/DynSUP/}.
\end{abstract}
\def\thefootnote{*}\footnotetext{These authors contributed equally to this work.}\def\thefootnote{\arabic{footnote}}
\section{Introduction}
\label{sec:intr}

Novel view synthesis, which aims to generate new views of a scene from a set of input images, is a fundamental computer vision task with widespread applications in virtual/augmented reality, robotics, and autonomous driving. Recent advances in neural rendering techniques like Neural Radiance Fields (NeRF)~\cite{mildenhall2020nerf} and 3D Gaussian Splatting (3D-GS)~\cite{kerbl20233d} have shown remarkable progress in achieving high-quality view synthesis. However, mainstream approaches typically rely on three restrictive assumptions: (1) the requirement of dense-view images, (2) known camera poses, and (3) static scene conditions. These assumptions significantly limit their practical applications in real-world scenarios where one or more conditions may not be satisfied.
\begin{figure}[!t]
\centering
    \includegraphics[width=\linewidth]{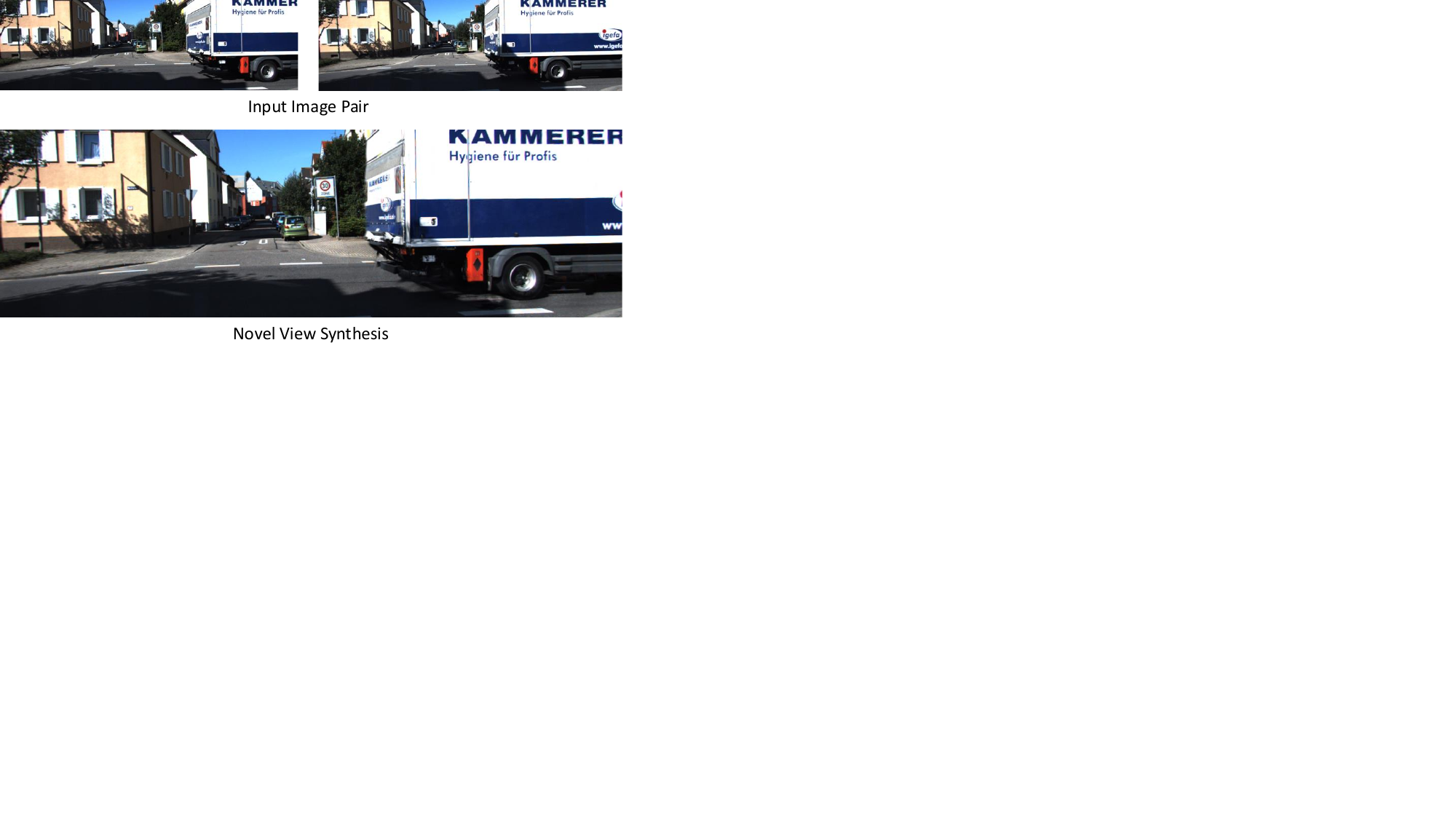}
\vspace{-1em}
\caption{\textbf{Dynamic Gaussian Splatting from An Unposed Image Pair. }Given two images captured at distinct moments with unknown poses in a dynamic environment, our method can fit dynamic Gaussian splatting and then synthesize a new image from a novel viewpoint at a different time.}
\label{fig:teaser}
\vspace{-1em}
\end{figure}
Recent works have relaxed some of these constraints. For example, to reduce the dependency on multiple images, methods like PixelSplat~\cite{charatan23pixelsplat} and MVSplat~\cite{chen2024mvsplat} leverage epipolar attention and cross-view transformer for reconstructing static scenes with sparse views. InstantSplat~\cite{fan2024instantsplat} further demonstrates a pose-free setting by integrating dense correspondence learning \cite{wang2024dust3r} to obtain dense point cloud with the relative pose, achieving promising results of static environments.

However, to the best of our knowledge, no existing method can handle dynamic scenes given sparse images without known poses. Dynamic scene reconstruction presents unique challenges due to relatively high-dimensional parameter space and complex geometric configurations. To address this challenge, we present DynSUP: Dynamic Gaussian Splatting from An Unposed Image Pair. This novel method enables high-quality dynamic scene reconstruction and rendering given two unposed images (see~\cref{fig:teaser}). Our approach consists of three key components. First, we introduce an object-level dense bundle adjustment, which decomposes dynamic scenes into piece-wise rigid components. By jointly optimizing reprojection loss with depth regularization loss, we achieve robust camera pose estimation and per-object motion recovery. Second, we develop $SE(3)$ Field-driven Gaussian Splatting where each Gaussian maintains an individual $SE(3)$ transformation initialized from object-level motion. This continuous transformation field enables fine-grained motion modeling while maintaining temporal consistency through regularization terms. These two components effectively bridge the geometric and photometric constraints. Additionally, we optimize the camera pose and per-object $SE(3)$ ratios for test image alignment. The main contributions of our work include:
\begin{itemize}
    \item We propose a novel method to fit Gaussian splats from an un-posed image pair in dynamic environments.
    \item We design a novel object-level dense bundle adjustment framework that can robustly recover 3D structure and motion of objects, which are then used to create $SE(3)$ Field-Driven 3D Gaussian splats, enabling novel view rendering of highly dynamic scenes.
    \item Extensive experiments on synthetic and real-world datasets demonstrate that our method significantly outperforms existing approaches designed for static scenes and/or known poses.
\end{itemize}

\section{Related Works} \label{sec:related_works}

\noindent\textbf{NVS with Sparse Views.}
Novel-view synthesis with sparse-view inputs has made significant progress through methods like FS-GS \cite{zhu2025fsgs}, DNSplatter \cite{turkulainen2024dn}  and DRGS~\cite{chung2024depth}, which leverage learned monocular depth or normal prior. InstantSplat \cite{fan2024instantsplat} uses dense stereo reconstruction model \cite{wang2024dust3r} for Gaussians initialization and can obtain fast 3DGS scene reconstruction. PixelSplat \cite{charatan23pixelsplat} and MVSplat \cite{chen2024mvsplat} utilize learned feature matching and epipolar constraints for sparse-view reconstruction. GRM~\cite{xu2024grm} and GS-LRM \cite{zhang2025gs} rely on large amounts of training data and resources to achieve few-shot reconstruction with transformer-based architecture. However, these approaches primarily target static scenes and struggle with dynamic content.

\noindent\textbf{Pose-free NVS.}
Typically, NeRF or 3DGS-based methods need accurate camera poses of input images, which are commonly obtained from the Structure-from-Motion algorithms~\cite{schonberger2016structure}. However, they often fail in the case of sparse-view inputs due to insufficient image correspondences. NeRFmm \cite{wang2021nerf} and BARF \cite{lin2021barf} propose to optimize coarse camera poses and NeRF jointly. PF-LRM \cite{wang2023pf} extends LRM \cite{hong2023lrm} to be applicable in pose-free scenes by using a differentiable PnP solver, but it shows limitation as it only focuses on the object-centric scene.  Nope-NeRF \cite{bian2023nope} and CF-GS \cite{Fu_2024_CVPR} leverage monocular depth estimation to constrain NeRF or 3DGS optimization, yet these pose-independent approaches generally presume the input are video sequences. DBARF~\cite{chen2023dbarf}, Flowcam~\cite{smith2023flowcam}, and CoPoNeRF \cite{hong2023unifying} try to integrate camera pose and radiance fields estimation in a single feed-forward pass. Recent work DUSt3R \cite{wang2024dust3r} and MASt3R~\cite{smart2024splatt3r} propose to regress the dense point maps of unposed input views in a global coordinate. Built upon these, Splatt3R \cite{hong2024pf3plat} predicts Gaussians from the frozen MASt3R backbone. GGRT \cite{li2024ggrt} designs a pose optimization network with a generalizable 3DGS model. 

\noindent\textbf{NVS in Dynamic Environments.}
Capturing dynamic scenes presents unique challenges due to fundamental ambiguity between camera and object motion. Prior work like 4D-GS \cite{4DGS} has addressed dynamic scene modeling but relies on known camera poses. Moreover, these approaches typically require multiple synchronized cameras or video sequences with known temporal ordering. Recent advances in urban scene modeling, including Street Gaussians \cite{yan2024street}, Driving Gaussians \cite{zhou2024drivinggaussian}, and $S^{3}$ Gaussian \cite{huang2024textit}, have proposed compositional frameworks that separately model static backgrounds and dynamic objects while optimizing a scene graph. However, these methods heavily rely on LiDAR point clouds for initialization and precise camera pose estimation. A notable recent breakthrough, Monst3R~\cite{zhang2024monst3r}, introduces a novel per-frame point-map estimation technique for dynamic scenes, enabling joint optimization of depth estimation, camera pose recovery and dense reconstruction from monocular video sequences. However, it can hardly achieve a photo-realistic rendering due to the point cloud-based 3D representation.

Overall, the existing methods cannot handle GS given sparse, un-posed images in dynamic environments. By contrast, our work bridges these domains by introducing the first method capable of handling dynamic scenes from just two unposed views through explicit SE(3) motion modeling and object-level bundle adjustment. Our method optimizes camera parameters and per-Gaussian motion fields to enable reliable dynamic GS.

\section{Problem Formulation} \label{sec:method}
\begin{figure*}[ht]
 \centering
    \includegraphics[width=\linewidth]{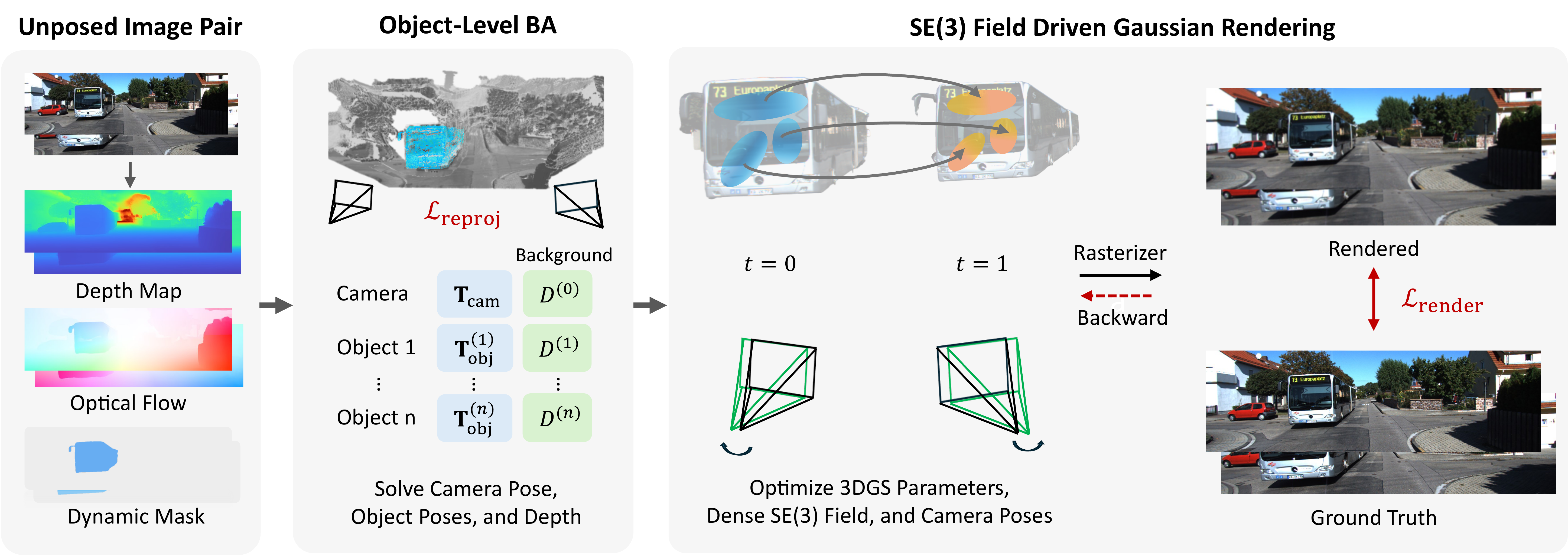}
    \vspace{-2em}
    \caption{\textbf{Overview of our DynSUP framework.} Given two unposed images, we first perform Object-level Dense Bundle Adjustment to estimate initial camera poses and object motions by decomposing the scene into piece-wise rigid components. The dense 3D Gaussian primitives are initialized with per-object $SE(3)$ transformations. In the $SE(3)$ Field-driven 3DGS stage, we jointly optimize the camera poses, per-Gaussian $SE(3)$ transformations, and Gaussian parameters to reconstruct the dynamic scene. The optimized $SE(3)$ field captures fine-grained motion details while maintaining temporal consistency. Finally, the dynamic scene is rendered using the optimized camera poses and $SE(3)$ field to generate high-quality novel-view synthesis results.}
    \label{fig:pipeline}
    \vspace{-1em}
\end{figure*}
\subsection{Overview}
We introduce a method for novel-view synthesis from two unposed images and intrinsic camera parameters using Gaussian Splatting. Our approach consists of two main stages: First, we employ Object-level Dense Bundle Adjustment (Sec. \ref{sec:dense_ba}) to reconstruct a dense point cloud and estimate rigid motions by decomposing the scene into piece-wise rigid components. Building upon the reconstructed geometry and initial object-level rigid motions, we develop an $SE(3)$ Field-driven Gaussian rendering framework (Sec. \ref{sec:se3_gs}) where each Gaussian maintains its individual $SE(3)$ transformation, enabling fine-grained dynamic motion modeling. We employ a differentiable pipeline to facilitate rendering in a dynamic setting where camera poses, per-Gaussian $SE(3)$ transformations, and Gaussian parameters are jointly optimized by minimizing the photometric loss. Fig. \ref{fig:pipeline} illustrates the overall pipeline of our Dynamic Two-views Pose-free GS.

\subsection{Preliminary} \label{subsec:preli}
3D Gaussians \cite{kerbl20233d} offers an explicit representation of a 3D scene using a collection of 3D Gaussians. Each 3D Gaussian is characterized by a mean point $X$ and a covariance matrix $\Sigma$, which together describe its shape and spread in space. The influence of a Gaussian on a point $X$ is modeled by the following form: $G(X)=\exp \left(-\frac{1}{2} X^T \Sigma^{-1} X\right)$. For optimization purposes, the covariance matrix $\Sigma$ can be decomposed into a rotation matrix $\mathbf{R}$ and a scaling matrix $\mathbf{S}$, allowing both the orientation and scale of each Gaussian to be learned during training: $\Sigma=\mathbf{R S S}^T \mathbf{R}^T$.
Differentiable splatting \cite{kerbl20233d} is used to project the 3D Gaussians onto the camera planes during novel view rendering. The projection involves a viewing transformation matrix $W$ and the Jacobian $J$ of the affine approximation to the projective transformation. The resulting covariance matrix $\Sigma^{\prime}$ in the camera's coordinate system can be computed as:$\Sigma^{\prime}=J W \Sigma W^T J^T$.

In summary, each 3D Gaussian is represented by position $X \in \mathbb{R}^3$, color described by spherical harmonic (SH) coefficients $\mathcal{C} \in \mathbb{R}^k$ (where $k$ is the number of SH functions used), opacity $\alpha \in \mathbb{R}$, rotation parameter $r \in \mathbb{R}^4$, and scaling factor $s \in \mathbb{R}^3$. For each pixel, the contributions of overlapping Gaussians are computed based on their color and opacity, with the total color being a weighted sum of individual Gaussian. The final blended color is calculated using:
$C=\sum_{i=1}^N c_i \alpha_i \prod_{j=1}^{i-1}\left(1-\alpha_j\right)
$, where $c_i$ and $\alpha_i$ represent the color and opacity of the $i$-th Gaussian, respectively.

\section{Method} \label{sec:method}
\subsection{Object-level Dense Bundle Adjustment} \label{sec:dense_ba}
Traditional Bundle Adjustment (BA) methods, including those used for 3D Gaussian initialization like COLMAP~\cite{schonberger2016structure}, are fundamentally limited to static scenes. These approaches can only optimize camera poses and 3D points by minimizing reprojection error under the assumption that the entire scene remains stationary. However, real-world scenarios rarely contain purely static structures - objects move, deform, and interact dynamically.

We introduce Object-level Dense Bundle Adjustment to address this limitation, which decomposes dynamic scenes into piecewise rigid components. Our key insight is that while the global scene may be dynamic, individual objects often exhibit local rigidity. By segmenting the scene into object-level components and treating each as a locally rigid structure, we can extend the power of traditional BA to dynamic scenarios.

Our approach first performs scene decomposition into object-level rigid components, followed by per-object dense BA optimization. This optimization framework jointly refines individual object poses and motions, static background structure, and camera parameters. The piecewise rigid formulation enables our method to optimize dense 3D point clouds reconstructed from two unposed images with known intrinsics while explicitly modeling object-level dynamics. We minimize reprojection error for each rigid component, along with depth regularization terms, through a fully differentiable optimization framework.

Let $I_{0}$ and $I_{1}$ denote two frames of a dynamic scene at timestamps $t = 0$ and $t = 1$, respectively, with camera intrinsic matrix $\mathbf{K}$. The initial depth maps $D_0$ and $D_1$ for these frames are obtained from monocular depth estimation~\cite{depth_anything_v2}. To establish pixel correspondences between $I_0$ and $I_1$, we leverage an optical flow network~\cite{xu2023unifying} to obtain forward optical flow $f_{0\rightarrow1}$ from $I_0$ to $I_1$. For a pixel $p$ and its homogeneous representation $\Tilde{p}$  with depth $D(p)$, the corresponding 3D point $P$ in the camera coordinate system of each frame is given by:
$P_0 = D_0(p_0)\mathbf{K}^{-1}\Tilde{p_0}$, $P_1 = D_1(p_1)\mathbf{K}^{-1}\Tilde{p_1}$.

Real-world dynamic scenes can often be decomposed into piecewise rigid components. Therefore, we first employ a dynamic object segmentation network~\cite{yang2021learning} to detect bounding boxes of dynamic objects. Then, we obtain precise motion segmentation masks by SAM-2~\cite{ravi2024sam2} with the detected boxes as prompts. These masks partition the pixels in each frame into distinct regions ${\mathcal{P}^{(0)}, \mathcal{P}^{(1)}, ..., \mathcal{P}^{(n)}}$, where $\mathcal{P}^{(0)}$ denotes the static background region and ${\mathcal{P}^{(1)}, ..., \mathcal{P}^{(n)}}$ represent dynamic object regions.

\noindent\textbf{Loss Function.}
For each region $\mathcal{P}^{(i)}$, we estimate a rigid transformation $\mathbf{T}^{(i)} \in S E(3)$ that describes the relative motion from $I_0$ to $I_1$ for that region. These transformations are initialized with PnP \cite{hartley2003multiple, lepetit2009ep} and RANSAC \cite{fischler1981random} using corresponding points from an optical flow network \cite{xu2023unifying}. For a 3D point $P_0$ corresponding to pixel $p_0$ in region $\mathcal{P}^{(i)}$ in frame $I_0$, its transformed position in $I_1$ is given by:$P_0^{\prime}=\mathbf{T}^{(i)} P_0.$

To ensure geometric consistency, we project $P_0^{\prime}$ onto the image plane of $I_1$, and it should ideally overlap with the tracked pixel position $p_1=p_0+\mathbf{f}_{0 \rightarrow 1}\left(p_0\right)$, where $\mathbf{f}_{0 \rightarrow 1}\left(p_0\right)$ is the optical flow from $I_0$ to $I_1$ at. However, dense optical flow can introduce noise, especially in scenarios of fast motion or severe occlusion. To mitigate this, we identify confident correspondences through a forward-backward consistency check \cite{meister2018unflow} using bidirectional flow \cite{xu2022gmflow}. The resulting confidence weight $\mathcal{W}_{\text {fwd}}$ indicates pixel areas with consistent forward $\mathbf{f}_{0 \rightarrow 1}$ and backward $\mathbf{f}_{1 \rightarrow 0}$ flows.
We formulate the reprojection loss for each pixel $p_0$ in region $\mathcal{P}^{(i)}$ is formulated as:
\begin{equation}
\small
\mathcal{L}_{\text {reproj}}=\sum_{p_0 \in \mathcal{P}^{(i)}} \mathcal{W}_{\text {fwd }}(p_0)\left\|\pi\left(\mathbf{K} \mathbf{T}^{(i)} \hat{P_0}\right)-p_1\right\|_1,
\label{equ:reproj_loss}
\end{equation}
where $\pi(\cdot)$ denotes the camera projection function, $\|\cdot\|_1$ denotes the L1 norm, $\mathcal{W}_{\text {fwd }}\left(p_0\right)$ is the confidence weight for pixel $p_0$ and $\hat{P_0}$ is the 3D point of pixel $p_0$ with optimized depth $\hat{D_0}$.

Dense bundle adjustment is sensitive to the quality of depth initialization. Without proper constraints, the optimized depths can deviate significantly from reasonable values. To address this issue, we introduce a depth regularization scheme that constrains the optimized depth $\hat{D}_0$ to approximate the initial monocular depth estimates $D_0$ through learnable scale $\theta$ and shift $\gamma$ parameters. Additionally, we bridge the estimated relative motions with the depth maps from two views. The transformed 3D point with optimized depth $\hat{{P}}_0^{\prime}=\mathbf{T}^{(i)}\hat{D_0}(p_0)\mathbf{K}^{-1}\tilde{p_0}$ should ideally match the scaled and shifted initial depth $D_1$ at the non-occluded tracked pixel $p_1$ in $I_1$ as $\hat{{P}}_1=\left(\theta_1 D_1(p_1)+\gamma_1\right) \mathbf{K}^{-1} \tilde{p}_1$. We formulate our depth regularization loss as follows:
\begin{equation}
% \small
\begin{aligned}
\mathcal{L}_{\mathrm{d}}= & \sum_{p_0 \in \mathcal{P}^{(i)}}\left\|\hat{D}_0\left(p_0\right)-\left(\theta_0 D_0\left(p_0\right)+\gamma_0\right)\right\|_1 \\
& +\sum_{p_0 \in \mathcal{P}^{(i)}} \mathcal{W}_{\mathrm{fwd}}(p_0)\left\|\hat{{P}}_0' - \hat{{P}}_1\right\|_1
\end{aligned} 
\end{equation}

The complete optimization objective combines reprojection and depth regularization terms:
\begin{equation}
\mathcal{L}_{ba} = \sum_{p \in \mathcal{P}^{(i)}}^n (\lambda_1 \mathcal{L}_{\text{reproj}} + \lambda_2 \mathcal{L}_{\text{d}})
\label{equ:ba_loss}
\end{equation}
where $\lambda_1$ and $\lambda_2$ are weighting parameters. We iteratively optimize this objective by adjusting the relative transformations $\mathbf{T}$ and depth values $\hat{D}$.

Through object-level BA, we solve for the relative motions $\mathbf{T}^{(i)}$ that describe the scene dynamics and per-object depth $D^{(i)}$. For the static region $\mathcal{P}^{(0)}$, the motion $\mathbf{T}^{(0)}$ represents the camera motion $\mathbf{T}_{\text {cam }}$. For dynamic regions $\mathcal{P}^{(i>0)}$, each $\mathbf{T}^{(i)}$ represents the combined motion of both the camera and the object. To recover the dynamic object motion in the world frame, we compute the object transformation $\mathbf{T}_{\text {obj }}^{(i)}$ as the inverse of the camera motion $\mathbf{T}_{\text {cam }}$, which is given by the motion of the static region $\mathbf{T}^{(0)}$, i.e., $\mathbf{T}_{\text {obj }}^{(i)}=\mathbf{T}_{\text {cam }}\left(\mathbf{T}^{(i)}\right)^{-1}$, for all dynamic regions $\mathcal{P}^{(i)}$ where $i \in[1, \ldots, n]$.

\noindent\textbf{Bidirectional Bundle Adjustment.}
Conventional Forward BA estimates the relative motion \( \mathbf{T} \) from \( I_0 \) to \( I_1 \) and reconstructs dense 3D points by minimizing the reprojection loss, as defined in \cref{equ:reproj_loss}. However, in challenging two-view scenarios, dense optical flow may suffer from inaccuracies, resulting in incorrect correspondences that adversely affect camera pose estimation. Furthermore, forward BA does not account for unseen regions in the subsequent frame, leading to potential information loss. To fully exploit the two-view information and ensure reliable correspondences, we introduce a Bidirectional BA approach, particularly focusing on the static region \( \mathcal{P}^{(0)} \).

We perform a forward-backward consistency check using bidirectional flow~\cite{meister2018unflow, xu2022gmflow}, which helps identify and discard unreliable correspondences due to occlusions or large motion. Confidence weights \( \mathcal{W}_{\text{fwd}} \) and \( \mathcal{W}_{\text{bwd}} \) are computed, indicating areas with consistent forward flow \( \mathbf{f}_{0 \to 1} \) and backward flow \( \mathbf{f}_{1 \to 0} \), respectively. By leveraging both directions, we extend the reprojection loss for the static region \( \mathcal{P}^{(0)} \).

For the forward BA, we compute the reprojection loss using the forward optical flow as described previously. In the Backward BA, we compute an additional reprojection loss by projecting the transformed 3D points from \( I_1 \) back onto the image plane of \( I_0 \) using the inverse camera transformation \( \mathbf{T}^{(0)^{-1}} \). We then compare these projections to the backward-tracked pixel location obtained via the backward optical flow \( \mathbf{f}_{1 \to 0} \), ensuring consistency across both frames.

Moreover, we extend the depth regularization to both frames, enforcing consistency between the 3D points reconstructed from \( I_0 \) and \( I_1 \). Aligning the depth values in this manner stabilizes the optimization and reduces the likelihood of depth discrepancies between frames.

\subsection{$SE(3)$ Field Driven Gaussian Rendering}\label{sec:se3_gs}
Building upon the Object-level Dense BA, we obtain dense 3D points in the world frame by projecting pixels from both frames using optimized depths and camera poses. For each pixel $p$, we initialize the 3D Gaussian primitive $\mathcal{G}$ with position, scale, and appearance parameters following~\cite{kerbl20233d}. Each Gaussian is also associated with its own rigid transformation $\mathbf{T} \in SE(3)$, forming a dense $SE(3)$ motion field regularized by the rigidity of the object. Compared to the strategy using the shared transformation, our method can improve the flexibility of Gaussian optimization and further enhance the rendering quality. Compared to prior work that models the motion implicitly with such as MLP~\cite{huang2023sc} or voxel grid~\cite{4DGS}, we use the $SE(3)$ field to explicitly represent deformation, which is more explainable and interpretable, allowing robust motion manipulation such as interpolation.

Concretely, to capture the motion of dynamic objects, we assign an initial transformation based on its corresponding object to each Gaussian. if a Gaussian corresponds to a 3D point from object $i$, its initial transformation $T$ is set to the object-level motion $\mathbf{T}_{\mathrm{obj}}^{(i)}$ recovered from the object BA. For static points belonging to the background region $\mathcal{P}^{(0)}$, we assign identity transformations as they remain stationary in the world frame.

To ensure stable optimization of rotations within the $SE(3)$ field, we adopt a continuous 6D rotation representation~\cite{zhou2019continuity} instead of quaternions, which is discontinuous in Euclidean space. 

Using these initial transformations and rotation representations, we utilize the $SE(3)$ field to drive the dynamic motion of Gaussians across frames. Specifically, in frame $I_0$, we can directly render the initial Gaussian $\mathcal{G}$ without applying any transformations. For frame $I_1$, each Gaussian undergoes its associated $SE(3)$ transformation according to its associated $SE(3)$ motion $\mathbf{T}$.
To achieve precise motion modeling, we refine the initial object-level transformations to a dense  $SE(3)$ field by allowing each Gaussian to optimize its individual transformation $\mathbf{T}_{\text{fine}} \in SE(3)$. The fine motion field is initialized from object-level transformations. If the Gaussian is initialized from the pixel from $\mathcal{P}^{(i)}$, then we initialize the $\mathbf{T}_{\text{fine}}$ with $\mathbf{T}_{\text{obj}}^{(i)}$.
Image rendering at different timestamps uses differentiable rasterizer $\Psi$~\cite{kerbl20233d}:
\begin{equation}
\hat{I}_t = \Psi(\mathcal{G}_i, \mathbf{T}_{\text{fine}}, \mathbf{T}_{\text{cam}}, \mathbf{K}),
\end{equation}
where $\mathcal{G}$ and $\mathbf{T}_{\text{fine}}$ denote the Gaussians and their fine transformations, $\mathbf{T}_{\text{cam}}$ is the camera motion, and $\mathbf{K}$ is the intrinsic camera matrix. The rendering loss combines $L_1$ loss and structural similarity loss $L_\text{D-SSIM}$ between the rendered image $\hat{I}_t$ and ground truth image $I_t$, following \cite{kerbl20233d}.
During optimization, we jointly refine Gaussian parameters, their associated $SE(3)$ transformations $\mathbf{T}_{\text{fine}}$, and camera poses $\mathbf{T}_{\text{cam}}$.

\noindent\textbf{$SE(3)$ Field Regularization.}
To promote smooth and physically plausible motion within each object region, we introduce regularization terms for both translation and rotation components of the $SE(3)$ field. The goal is to minimize the variance of transformations within each dynamic region, ensuring coherent motion for Gaussian primitives belonging to the same object.

For translation regularization $\mathcal{L}_{t_i}$, we compute the average translation \( \mathbf{v}_{t_i} \) for each dynamic region \( P_i \) and penalize deviations from this average using the Huber loss \( \mathcal{H} \). Similarly, for rotation $\mathcal{L}_{r_i}$, we compute the average rotation \( \mathbf{v}_{r_i} \) in 6D space and regularize using the Huber loss applied to the normalized rotation representation \( \mathcal{N}(\mathbf{r}_j) \) of each Gaussian primitive.

The total regularization loss is the weighted sum of translation and rotation terms:
\begin{equation}
\mathcal{L}_{\text{reg}} = \lambda_t \sum_{i > 0} \mathcal{L}_{t_i} + \lambda_r \sum_{i > 0} \mathcal{L}_{r_i},
\label{equ:se3_reg}
\end{equation}
where \( \lambda_t \) and \( \lambda_r \) (set to 1 by default) control the weighting for translation and rotation regularization, respectively. The summation runs over all dynamic regions \( \mathcal{P}^{(i)} \) excluding the static background region \( \mathcal{P}^{(0)} \).

\begin{table*}[t]
\setlength{\tabcolsep}{8pt} 
\centering
% \footnotesize
\small
\caption{\textbf{Novel-view synthesis results on KITTI \cite{Geiger2012CVPR} and Kubric \cite{greff2021kubric} datasets}. The best results are highlighted in bold. Our method shows consistent superior performance on both datasets.}
\label{table:nvs}
\begin{tabular}{cccccccc}
\toprule
& \multicolumn{3}{c}{KITTI~\cite{Geiger2012CVPR}} & & \multicolumn{3}{c}{Kubric~\cite{greff2021kubric}} \\
\cline{2-4} \cline{6-8}
Methods & PSNR $\uparrow$ & SSIM $\uparrow$ & LPIPS $\downarrow$ & & PSNR $\uparrow$ & SSIM $\uparrow$ & LPIPS $\downarrow$ \\
\midrule
4DGS~\cite{4DGS} & 17.97 & 0.58 & 0.36 & & 20.33 & 0.67 & 0.39 \\
SC-GS~\cite{huang2023sc} & 18.81 & 0.58 & 0.28 & & 22.54 & 0.74 & 0.22 \\
InstantSplat~\cite{fan2024instantsplat} & 22.11 & 0.79 & 0.19 & & 25.80 & 0.91 & 0.10 \\
Ours & \textbf{24.71} & \textbf{0.82} & \textbf{0.13} & & \textbf{33.86} & \textbf{0.97} & \textbf{0.03} \\
\bottomrule
\end{tabular}
\vspace{-1em}
\end{table*}

\subsection{Test-time Poses and $SE(3)$ Ratios Alignment}
Unlike conventional approaches where exact camera poses for test views are known and estimated alongside the training views~\cite{mildenhall2020nerf,kerbl20233d}, our scenario involves test views with unknown poses. Following strategies from prior work~\cite{wang2021nerf, fan2024instantsplat}, we fix the Gaussian Splatting model trained on training views and optimize the camera poses for test views during inference. 

Additionally, we introduce an additional optimization step to align the temporal positions of dynamic objects in the test images by adjusting the interpolation ratios of their  $SE(3)$ transformations. We initialize the interpolation ratio for each object's $SE(3)$ transformation to 0.5, assuming the dynamic objects in the test image are temporally situated halfway between two training images. During optimization, these per-object interpolation factors are refined to their optimal values, effectively aligning the dynamic object motions with the test views. The optimization objective minimizes the photometric discrepancy between the synthesized and actual test images, jointly optimizing the camera poses and dynamic object motions for precise rendering. The rendering process for the test view is defined as follows:
\begin{equation}
    \hat{I}_\text{test} = \Psi(G, \mathbf{T}_{\text{fine}}, \mathbf{T}_{\text{cam}}, \mathbf{K}, r_{\text{obj}})
\end{equation}
where $\hat{I}_\text{test}$ is the synthesized test image, $G$ represents the Gaussian primitives, $\mathbf{T}_{\text{fine}}$ denotes the fine-grained $SE(3)$ transformations, $\mathbf{T}_{\text{cam}}$ is the camera pose, $\mathbf{K}$ is the camera intrinsic matrix, and $r_{\text{obj}}$ represents the per-object $SE(3)$ ratios. In this formulation, the per-object interpolation factors $r_{\text {obj }}$ adjust the $S E(3)$ transformations of dynamic objects by interpolating between their transformations at the training timestamps. The optimization objective minimizes the photometric discrepancy between the rendered image $\hat{I}_{\text {test }}$ and the actual test image $I_{\text {test }}$ using the loss following \cite{kerbl20233d}, jointly refining the camera poses and dynamic object motions to achieve accurate rendering.

\noindent\textbf{Implementation Details.} \label{subsec:implementation}
For Object-level Dense BA, we use Depth Anything V2~\cite{depth_anything_v2} for monocular depth estimation and GMFlow~\cite{xu2023unifying} for optical flow. Rigid motion detection is handled using Rigidmask~\cite{yang2021learning} refined by SAM-2~\cite{ravi2024sam2}. The bundle adjustment is implemented in PyTorch with the Adam optimizer, parameterized using PyPose~\cite{wang2023pypose} for pose parameters. We set the learning rates to 1e-3 for depth and 1e-4 for pose optimization, with loss weights $\lambda_1 = 1.0$ for reprojection and $\lambda_2 = 0.1$ for depth regularization, running for 2000 iterations per image pair. For $SE(3)$ Field-driven Gaussian Splatting, we optimize with the Adam optimizer for 1000 iterations. We conduct our experiments on an NVIDIA RTX 4090 GPU.

\begin{figure*}[t]
 \centering
\includegraphics[width=\linewidth]{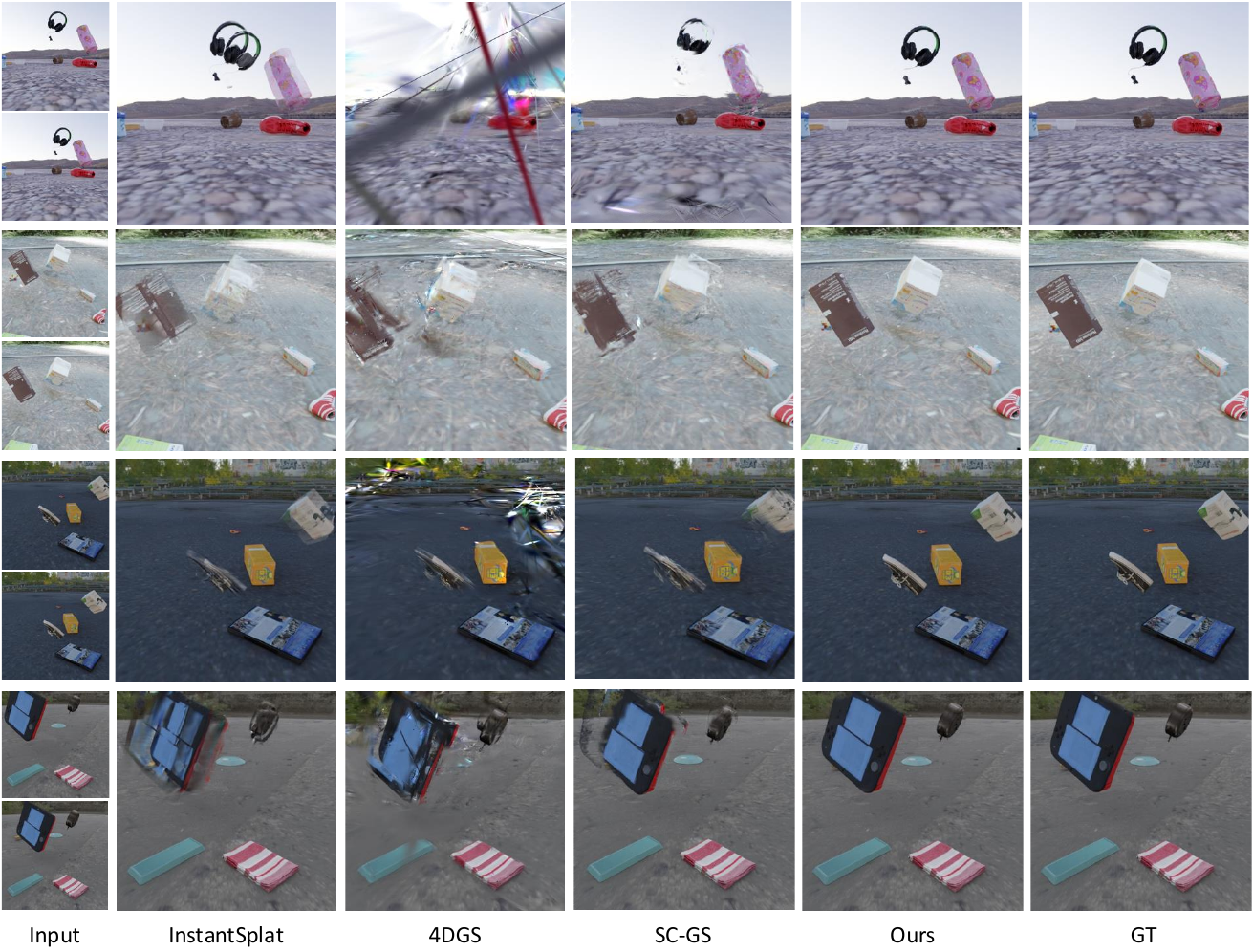}
    \caption{\textbf{Qualitative comparison on the Kubric dataset~\cite{greff2021kubric}.} Our method produces high-fidelity results for challenging scenes with multiple fast-moving objects.}
    \label{fig:sota_kubric}
\vspace{-1em}
\end{figure*}

\begin{figure*}[t]
 \centering
    \includegraphics[width=\linewidth]{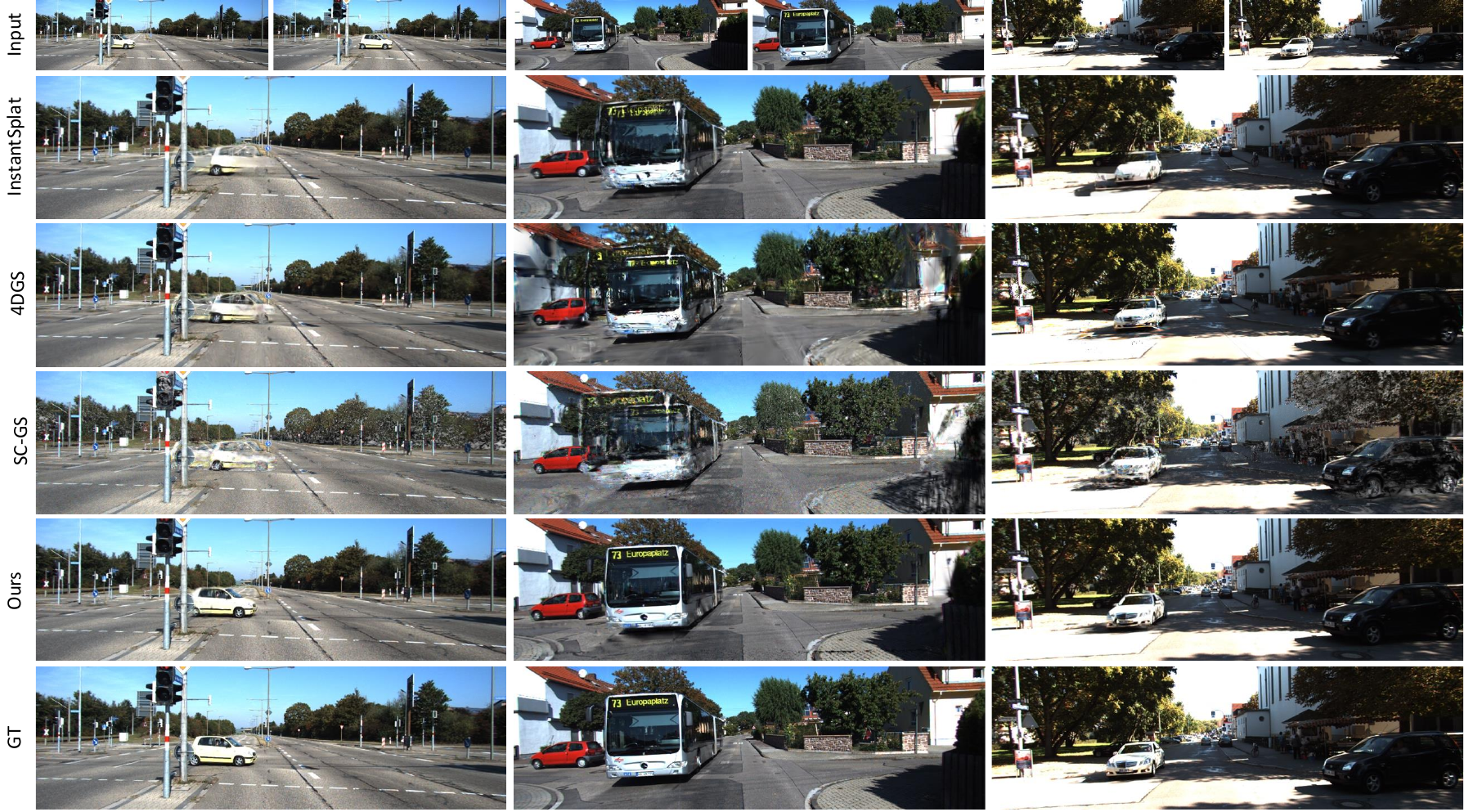}
    \caption{\textbf{Qualitative comparison on the KITTI dataset~\cite{Geiger2012CVPR}}. Our method handles complex urban environments with varying object and camera motion better than baseline approaches.}
    \label{fig:sota_kitti}
\vspace{-1em}
\end{figure*}

\section{Experiment} \label{sec:experiment}

\textbf{Datasets.} In our setup, for three consecutive images in a monocular video, we take the first and third frame as a pair for training and the intermediate frame for evaluation. We conduct extensive experiments on real-world and synthetic datasets.
The \textbf{KITTI} dataset~\cite{Geiger2012CVPR} provides real-world driving scenarios captured by vehicle-mounted cameras, featuring complex urban environments with dynamic elements such as vehicles and pedestrians. We evaluate our method using 180 image pairs from 24 scenes, representing diverse motion levels, including static, slow, and fast camera and object movements.
The \textbf{Kubric} dataset~\cite{greff2021kubric} offers synthetic sequences with precise camera parameters and photorealistic images suitable for quantitative evaluation. We generate 100 image pairs from 39 sequences, covering multiple moving rigid objects with varying trajectories and complex lighting.

\noindent\textbf{Baselines.} We compare our approach with state-of-the-art methods, featuring pose-free input, dynamic scene modeling, and sparse deformation representation.
\textbf{InstantSplat} \cite{fan2024instantsplat} is a pose-free method for novel-view synthesis of static scenes. It can handle sparse views with the dense point map predicted by DUSt3R~\cite{wang2024dust3r} as an initialization.
\textbf{4DGS} \cite{4DGS} can model the radiance field of a dynamic scene with a deformation field represented by a 4D grid and a tiny MLP. It is originally designed for dense-view videos with camera poses.
\textbf{SC-GS }\cite{huang2023sc} relies on sparse control points learned by an MLP over time to capture scene dynamics. It features smoother and more locally consistent motion. But it also requires dense-view videos with camera poses.

Note that 4DGS and SC-GS originally rely on COLMAP~\cite{schonberger2016structure} for camera poses and an initial point cloud. However, COLMAP struggles to estimate poses in our setting, only two-view observation of a dynamic scene and sometimes little to no disparity in the static background. Therefore, we use the camera poses and dense point map predicted by DUSt3R~\cite{wang2024dust3r} for experiments with 4DGS and SC-GS.

\noindent\textbf{Metrics.} We evaluate our method and baselines with standard metrics for novel-view synthesis: PSNR (Peak Signal-to-Noise Ratio) measures pixel fidelity, SSIM (Structural Similarity Index Measure \cite{wang2004image}) quantifies structural similarity, and LPIPS (Learned Perceptual Image Patch Similarity \cite{zhang2018unreasonable}) captures perceptual similarity, assessing visual realism.

\subsection{Comparisons with State-of-the-art Methods} \label{subsec:comp_sota}

We show qualitative comparison with state-of-the-art methods in \cref{fig:sota_kitti,fig:sota_kubric}. 
InstantSplat~\cite{fan2024instantsplat} achieves great visual results in static regions but produces replicas of the same object at different places due to lack of motion modeling. 
The grid-based deformation field of 4D-GS~\cite{4DGS} has a large capacity for complex motion but performs poorly when the number of observation in space and time is limited, producing floaters in novel views and temporally inconsistent results at the intermediate timestamp for evaluation.
SC-GS~\cite{huang2023sc} enhances local motion consistency by incorporating sparse control points rather than modeling per-point movement. However, despite these improvements, SC-GS still struggles to accurately recover motion in highly dynamic scenes, as the optimizable control points are still highly ambiguous with two-view observations and difficult to align with real objects.
Benefiting from the dense object-level BA that solves for a single transformation with all points on each object, our method is robust to limited observations and large motion. The per-Gaussian $SE(3)$ transformation and camera pose optimization followed by ratio alignment further contribute to high-quality rendering of both dynamic objects and the static background, which closely match the ground truth. In accordance with the qualitative comparison, the quantitative evaluation in \cref{table:nvs} shows consistently superior performance of our method than state-of-the-art methods on novel-view synthesis metrics on both datasets. 

\subsection{Ablation Study} 
\label{subsec:ablation}

\begin{figure}[t]
 \centering
    \includegraphics[width=\linewidth]{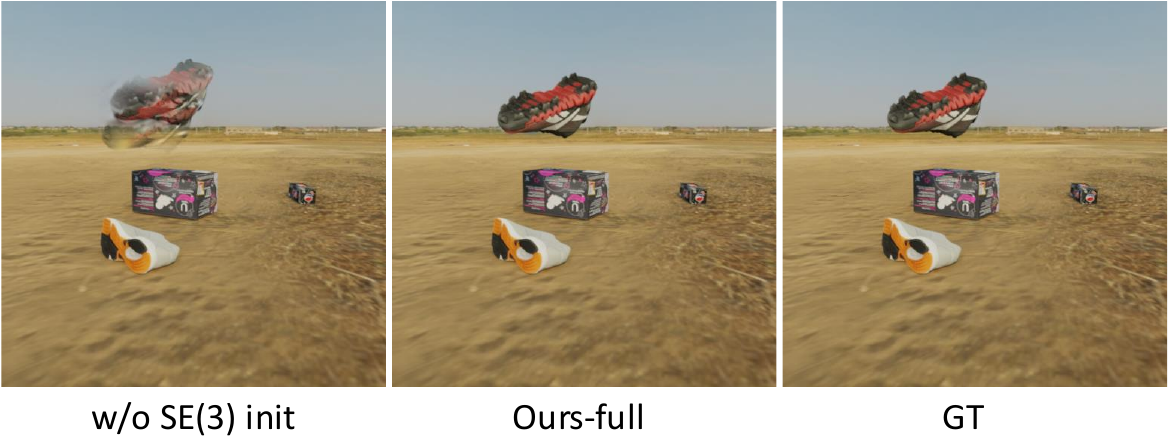}
\caption{\textbf{Ablation study on the Kubric dataset~\cite{greff2021kubric} for $SE(3)$ initialization.}}
\label{fig:ablation}
\vspace{-1em}
\end{figure}

We conduct ablation study on the Kubric dataset \cite{greff2021kubric} to evaluate the importance of two key components: $SE(3)$ motion initialization and test-time ratio optimization. 

As shown in the first line of \cref{tab:ablation}, replacing object-level BA for SE(3) initialization by identity transformations results in significantly worse performance. The optimization fails to converge without proper initialization, particularly in regions with large motion. This highlights the importance of motion estimation on the object level for highly dynamic scenes. Qualitative results in \cref{fig:ablation} further demonstrate that dynamic regions appear blurry without $SE(3)$ initialization. 

We validate the impact of optimizing per-object $SE(3)$ interpolation ratios during test time by the second line of \cref{tab:ablation}. Fixing the interpolation ratio to 0.5 leads to reasonable but inaccurate estimation of the motion, leading to a slight performance drop. 

With motion initialization and ratio optimization, our full model outperforms all variants and ensures high-quality rendering with consistent object motion, confirming that $SE(3)$ motion initialization and test-time ratio optimization are critical for accurate and consistent dynamic scene reconstruction from sparse views.

\begin{table}[ht]
\footnotesize
\centering
\caption{\textbf{Ablation study on the Kubric dataset~\cite{greff2021kubric}.} $SE(3)$ initialization is crucial and test-time ratio alignment further improves the performance.}
\label{tab:ablation}
\resizebox{\columnwidth}{!}{
\begin{tabular}{@{}lccc@{}}
\toprule
\multicolumn{1}{c}{Method}   & PSNR  $\uparrow$          & SSIM   $\uparrow$        & LPIPS    $\uparrow$      \\
\midrule
w/o $SE(3)$ initialization            & 26.00          & 0.92          & 0.09          \\
w/o test-time ratio alignment & 32.14          & 0.95          & 0.04          \\
Ours                         & \textbf{33.86} & \textbf{0.97} & \textbf{0.03} \\
\bottomrule
\end{tabular}
}
\end{table}

\section{Conclusion} \label{sec:conclusion}
We introduced a novel approach that integrates Dense Bundle Adjustment with 3D Gaussian Splatting for dynamic scene reconstruction using dense $SE(3)$ fields from two views. Our method explicitly recovers camera poses and object motions through object-level Dense BA, facilitating accurate initialization of 3D Gaussians with dynamic motion. By incorporating a dense $SE(3)$ field, we enable each Gaussian to optimize its individual transformation while maintaining motion consistency through our regularization scheme. Despite its strengths, our approach has limitations. It is primarily designed for rigid or piecewise rigid motions and struggles with non-rigid deformations, where the assumption of rigid transformations no longer holds. Additionally, the accuracy of our method depends heavily on accurate initial motion segmentation. Coarse or inaccurate segmentation of dynamic objects can negatively impact motion estimation and subsequent rendering quality.
{
    \small
    \bibliographystyle{ieeenat_fullname}
    \bibliography{main}
}

\end{document}